\documentclass[conference]{IEEEtran}
\IEEEoverridecommandlockouts
\usepackage{cite}
\usepackage{amsmath,amssymb,amsfonts}
\usepackage{algorithmic}
\usepackage{graphicx}
\usepackage{graphics}
\usepackage{textcomp}
\usepackage{xcolor}

\usepackage{booktabs} 
\usepackage{subfig} 
\usepackage{multirow} 
\usepackage{hyperref}

\def\BibTeX{{\rm B\kern-.05em{\sc i\kern-.025em b}\kern-.08em
    T\kern-.1667em\lower.7ex\hbox{E}\kern-.125emX}}

\begin{document}
\title{Improved Kidney Stone Recognition Through Attention and Multi-View Feature Fusion Strategies 
}


\author{Elias Villalvazo-Avila$^{1,2}$, Francisco Lopez-Tiro$^{1,2,3}$, Jonathan El-Beze$^{4}$, Jacques Hubert$^{4}$, \\ Miguel Gonzalez-Mendoza*$^{,1}$, Gilberto Ochoa-Ruiz*$^{,1,2}$, Christian Daul*$^{,3}$ \\

\thanks{$^{1}$Tecnologico de Monterrey, School of Sciences and Engineering, Mexico}%
\thanks{$^{2}$CV-INSIDE Lab Member, Mexico}
\thanks{$^{3}$CRAN (UMR 7039, Université de Lorraine and CNRS), Nancy, France}%
\thanks{$^{4}$CHRU Nancy, Service d’urologie de Brabois, Vand{\oe}ure-l\`es-Nancy, France}%
\thanks{*Corresponding authors:} %
\thanks{gilberto.ochoa@tec.mx, christian.daul@univ-lorraine.fr}
}

\maketitle

\begin{abstract}This contribution presents a deep learning method for the extraction and fusion of information relating to kidney stone fragments acquired from different viewpoints of the endoscope. Surface and section fragment images are jointly used during the training of the classifier to improve the discrimination power of the features by adding attention layers at the end of each convolutional block. This approach is specifically designed to mimic the  morpho-constitutional analysis performed in ex-vivo by biologists  to visually identify kidney stones by inspecting both views. The addition of attention mechanisms to the backbone improved the results of single view extraction backbones by 4\% on average. Moreover, in comparison to the state-of-the-art, the fusion of the deep features improved the overall results up to 11\% in terms of kidney stone classification accuracy.

\end{abstract}

\begin{IEEEkeywords}
Multi-view, multimodal classification, attention, CBAM, ureteroscopy, deep learning.
\end{IEEEkeywords}

\section{Introduction} 
Urolithiasis refers to the formation of stony concretions in the bladder or urinary tract \cite{hall2009nephrolithiasis, kasidas2004renal}. It represents a major public health issue in industrialized countries: at least 10\% of the population appears to have a kidney stone and the risk of inappropriate treatment due to an incorrect stone type identification  can concern up to 40\% of patients\cite{kartha2013impact, scales2012prevalence}.
%
Therefore, the development of novel diagnosis and characterization tools for assisting clinicians is strongly encouraged by the urology community \cite{daudon2004clinical, estrade2017should}. Indeed, the in-vivo recognition of the type of kidney stones is an important aspect in the diagnosis, as it allows to prescribe adequate and personalized treatments in order to avoid relapses \cite{friedlander2015diet, kartha2013impact, viljoen2019renal}. 

The morpho-constitutional analysis (MCA) developed by Daudon et al. \cite{daudon2016comprehensive} is the reference method for the ex-vivo identification of kidney stones which were fragmented and extracted during an ureteroscopy. MCA is performed by biologists working in a laboratory, and consists of two complementary analyses. A Fourier transform infrared spectroscopy (FTIR) analysis provides the chemical composition of the kidney stone,  
whereas  a visual inspection of the fragment observed with a microscope  allows for the description of the crystalline structure based on colors and textures   \cite{corrales2021classification}. Both the FTIR analysis and a rigorous visual inspection of the fragment surface and section are required to unequivocally identify the kidney stone type.

However, fragmenting kidney stones with a laser and extracting them from the kidneys and ureters is a tedious procedure lasting between 30 and 60 minutes. Lasers can also be used to vaporize the fragments. Such dusting procedures significantly speed-up ureteroscopies and diminishes the infection risks, with the major drawback that MCA analyses become impossible. To overcome this issue, kidney stones can be visually identified on a screen by few experts \cite{estrade2017should}. Becoming such an expert entails extensive training, making their incorporation in the clinical practice unfeasible. Moreover, this visual  kidney stone recognition by urologists is operator dependent. AI techniques assessing endoscopic images could lead to an automated and operator independent in-vivo recognition.  


\begin{figure}[]
\centering
\includegraphics[width=0.99 \linewidth]{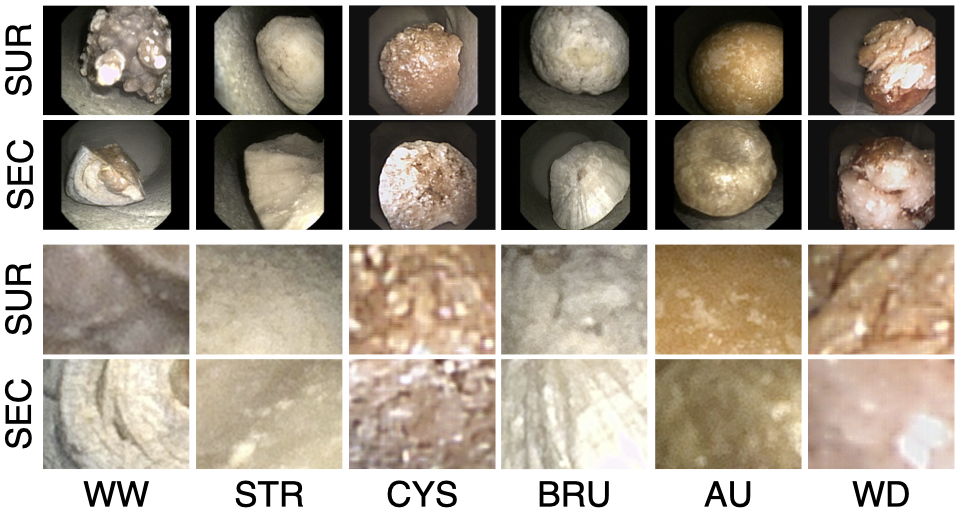}
\caption{Examples of kidney stone images of the used dataset \cite{el2022evaluation}. The latter consists of the six most common kidney stone types, namely whewellite (WW), weddellite (WD),  uric acid (AU),  struvite (STR), brushite (BRU), and cystine (CYS).} \label{fig:dataset}
\end{figure}

Despite the importance of this problem, only few works \cite{martinez2020towards, lopez2021assessing, ochoa2022vivo} have dealt with the identification  of kidney stones seen in images acquired with an ureteroscope.
However, none of these works have introduced a mechanism for fusing information (i.e., feature maps) of the section and surface views of a given kidney stone, which is what specialists do in clinical practice.
%
As noticeable in the two upper endoscopic image rows of Fig.~\ref{fig:dataset}, the aspect of the surface (SUR) and section (SEC) of kidney stone fragments depends on the urinary stone type. Existing methods have trained classifiers using features extracted from each image type, without taking into account the practices described by Daudon using the MCA analysis.
%

%
This contribution takes inspiration from recent works in multi-view fusion strategies \cite{geras2017high, seeland2021multi, sleeman2021multimodal}, which seek to combine characteristics from different sources or modalities to further improve 
machine learning based classification models.
The aim of combining/fusing the features extracted from surface and section images is to increase the amount of discriminant information to improve the accuracy of the classification.     
This approach based on feature fusion  is also extended with attention mechanisms to further improve the classification performance  (via feature refinement through attention).


The rest of this paper is organized as follows. Section \ref{sota} describes previous works dealing with the identification of kidney stones. Section \ref{mandm} starts with the description of  the data used in this contribution. Then, this section presents a novel  kidney stone   classification approach based on attention and fusion, and ends with the training step of the proposed deep-learning (DL) model. Section \ref{res} discusses the obtained results, while Section \ref{conclusion}  concludes the article.

\section{State-of-the-Art}
\label{sota}

The first works \cite{serrat2017mystone, martinez2020towards} dealing with the classification of kidney stones were based on shallow machine learning (SML) approaches, i.e., they used expert knowledge during the feature extraction. For instance, in \cite{martinez2020towards}, texture (local binary pattern histograms) and color (values in the hue/saturation/intensity space) information were gathered in feature vectors and treated by a random forest classifier to identify four kidney stone types. The results showed that using data from both section and surface images can lead to promising results. However, further work of these authors \cite{lopez2021assessing} has shown that SML-methods under-perform when compared with DL-methods in the context of kidney stone classification. 

In recent works \cite{ochoa2022vivo, black2020deep}, the performance was effectively improved by using DL-based methods. Encouraging results showed the potential of CNNs to extract sufficiently discriminative features in surface and section images. In addition, it was also shown that training a neural network by combining information extracted from both section and surface images improves the performance of the models. Furthermore, the results in  \cite{lopez2022boosting} showed that training models in different image distributions also improves the classification performance.
However, among all these solutions, not a single approach has been proposed to fuse the information extracted from images of the surface and section of urinary stone fragments.
Therefore, this contribution proposes a DL-based method that extracts and fuses information from both images types to assess whether image fusion can lead to an improvement of the classification performances in this task.

\section{Materials and methods}
\label{mandm}
\subsection{Dataset} 
\label{datasets}

%
%

The dataset used in this contribution was built for kidney stone fragments whose types were determined during MCA, i.e., the data were annotated using the reference laboratory procedure \cite{daudon2016comprehensive}.     
Images were acquired with an ureteroscope by placing the fragments inside a tubular shaped enclosure having a diameter and a color close to that of the ureters and their internal epithelial wall, respectively. As detailed in \cite{el2022evaluation}, although the images were acquired in ex-vivo, they are quite realistic since the environment and the illumination are very close to those observed in in-vivo, whereas the acquisitions were made with an endoscope and a light source actually used during an ureteroscopy. Table \ref{tab:dataset} shows that the dataset consists of 246 and 163  surface and section images, respectively.

\begin{table}[]
\centering
\caption{Endoscopic dataset. Number of images per class. SUR and SEC views contain 1000 patches per class each, while the MIX (SUR+SEC) contains 2000 patches per class.}
\vspace{-0.15cm}
\label{tab:dataset}
\begin{tabular}{@{}ccccc@{}}
\toprule
Type & Main component &  Surface & Section & MIX \\ \midrule  \vspace{-0.05cm}
Ia & Whewellite (WW) & 62  & 25   & 87  \\  \vspace{-0.05cm}
IIa & Weddellite (WD) & 13  & 12  & 25   \\  \vspace{-0.05cm}
IIIa & Acide Urique (AU) & 58  & 50  & 108  \\  \vspace{-0.05cm}
IVc & Struvite (STR) & 43  & 24  & 67  \\  \vspace{-0.05cm}
IVd & Brushite (BRU) & 23  & 4   & 27  \\  \vspace{-0.05cm}
Va & Cystine (CYS) & 47   & 48   & 95   \\ \cmidrule(l){2-5}   \vspace{-0.05cm}
 &  All types & 246   & 163  & 409   \\ \bottomrule  \vspace{-0.05cm}
\end{tabular}
\end{table}

As noticeable in Table \ref{tab:dataset}, rather few images are available for the six kidney stone types and the classes are imbalanced. It has been shown in previous works \cite{lopez2021assessing, martinez2020towards, ochoa2022vivo} that extracting from the images square patches with a maximal overlap of 20 pixels and with an appropriate size allows for capturing non redundant information including locally representative color and texture data.   
In theses works, the square patch side length was a hyper-parameter whose optimal value of 256 pixels was adjusted in the test phase. In this contribution, the patch size is also $256 \times 256$ pixels (see the two last rows of Fig.~\ref{fig:dataset}). Extracting patches from  the images (which are also whitened, see \cite{martinez2020towards}) and performing data augmentation are two means to increase the amount of data and to balance the classes. 

%
%


\subsection{Proposed Approach}
\textbf{Multi-View Classification. }
Multi-View (MV) classification seeks to combine characteristics from different sources (here the image types). The accuracy of object identification increases due to  the diversity of the features extracted from different sources and that are fused \cite{li2016multi, sleeman2021multimodal}. Thus, the performance of a DL-model can be improved by optimizing multiple functions, one per each image type. MV-fusion in CNNs has a particular interest when images from a single source do not carry sufficiently discriminative information for performing an accurate classification.

\textbf{Attention. }
CNNs have demonstrated their capability to solve a variety of visual tasks, such as classification. 
However, the reasoning for performing a classification task is often  unintelligible (i.e., the model can be seen as a black-box) limiting the understanding of the modle inner workings. One approach to visualize and to improve the representation power of the CNNs lies in the use of attention layers, which are scalar matrices representing the relative importance of a given layer activation at different locations with respect to the target task \cite{jetley2018learn, woo2018cbam}. By using attention, the model can focus on the important features, while suppressing the unnecessary ones.
%

\begin{figure*}[]
\includegraphics[width=.95\linewidth]{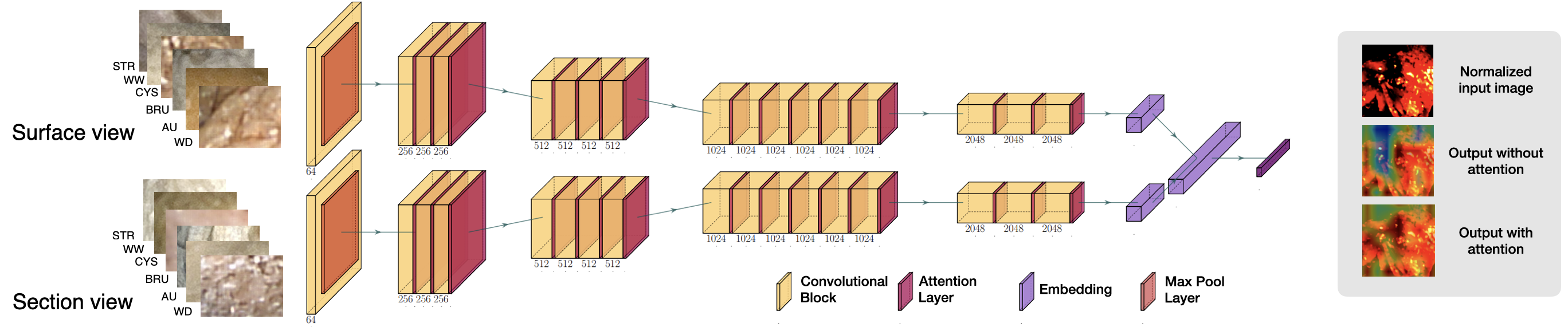}
\centering
\vspace{-0.15cm}
\caption{Proposed Multi-View model with attention. The first part of the model corresponds to the duplicated feature extraction layers from the ResNet50 model. These layers are followed by the fusion layer, which combines information from the two views (i.e., from the two image types). The fused feature map is then connected to the classification layer.}
\label{fig:mv_figure}
\end{figure*}

\textbf{Convolutional Block Attention Module. }
A recent attempt to incorporate attention into CNNs to improve their performance was described in \cite{woo2018cbam}. The Convolutional Block Attention Module (CBAM) consists of two attention sub-modules, namely i) a channel and ii) a spatial attention which are applied in that sequential order. Channel attention aims to focus on  feature maps that are important for the learning step and enhances their relevance. On the other hand, spatial attention attempts to learn more discriminant points in the feature maps. 
Combining both feature maps has demonstrated to yield an improvement in classification performance.

\textbf{Proposed Model. }
This contribution uses a ResNet50-based model  pre-trained with ImageNet for performing the classification of the kidney stone types listed in Table \ref{tab:dataset}. In a preliminary experiment, it was observed that initializing the network  on ImageNet improves the classification results, even if the distribution of this natural image dataset differs from that of the endoscopic image dataset \cite{lopez2022boosting}.
Then, an additional attention module consisting of the sequential application of channel and a spatial attention layers were added at the end of each conv block, as proposed in \cite{woo2018cbam}. For ResNet50, a total of 16 attention layers were added (see Fig \ref{fig:mv_figure}). 
Finally, the baseline model and the modified version with attention were used to train the MV fusion models without, and with attention, respectively. To fuse the features, two late-fusion strategies are explored: feature concatenation, and max-pooling of the individual features obtained by each view response \cite{sleeman2021multimodal, seeland2021multi}.

ResNet50 was selected as the base model since this architecture produced the best performance on the three different views from the state-of-the-art (surface, section, and both image types mixed) \cite{lopez2021assessing, lopez2022boosting}. For this model, CBAM blocks were added at the end of every convolutional block, as suggested in \cite{woo2018cbam}. The purpose of having attention added at multiple points in the network is to get subsequent refined feature maps from the intermediate feature maps.
The selected fusion mechanism is late fusion, in which the feature extraction and the learning is done independently before the final classification. The two approaches used for effectively fusing data  either concatenate  the deep features or merge the features by applying max-pooling. Finally, this fused features are used for the classification task.
Moreover, we aim to create a model that simulates how MCA is carried out, by combining information of surface and section views in the same model, which is more alike to what specialists do in clinical practice.

\subsection{Training}
\label{training}
\textbf{Single-view Model.} 
The base model used was evaluated in three scenarios: using of only surface features, using only of section patches, or  by combining  both views.
The model was trained for 30 epochs using a batch size of 32, along with the Adam optimizer with a learning rate of $2e^{-4}$. Finally, the representations obtained from the model are passed to fully connected layers with 512, 256, and 6 neurons each, with ReLU as activation function, batch normalization, and a dropout probability of 0.5.
The mixed views  are used to train a base model for the creation of the MV model.

\textbf{Multi-View Model.} 
For the training of the MV-fusion model, the feature extraction layers from the single-view model are frozen and then duplicated. One head processes the patches of the section view, wheareas the second head treats the surface patches. 
These layers are followed by the fusion layer, which mixes the information of  both views. The first fusion strategy consists of a stack of feature vectors on which max-pooling is applied. 
The second fusion method concatenates the features obtained by each view. 
Finally, the resulting representations are connected to a sequence of FC layers. 

The proposed model is shown in Fig. \ref{fig:mv_figure}. Since feature extraction layers are frozen, any difference in the performance lies on the fusion mechanism and the FC layers.

\begin{table*}[t!]
\centering
\caption{Mean $\pm$ standard deviation assessed for four quality criteria. Each model was executed five times.} 
\vspace{-0.15cm}
\label{tab:results1}
\begin{tabular}{@{}cccccl@{}}
\toprule  \vspace{-0.03cm}
View & Accuracy & Precision & Recall & F1-score & \multicolumn{1}{c}{Model description} \\ \midrule \vspace{-0.03cm}
\multirow{2}{*}{Surface} & 0.856 $\pm$ 0.030 & 0.872 $\pm$ 0.022 & 0.856 $\pm$ 0.030 & 0.858 $\pm$ 0.034 & Base model \\ \vspace{-0.03cm}
 & 0.888 $\pm$ 0.028 & 0.896 $\pm$ 0.024 & 0.886 $\pm$ 0.026 & 0.886 $\pm$ 0.026 & Base model + Attention \\ \midrule \vspace{-0.03cm}
\multirow{2}{*}{Section} & 0.836 $\pm$ 0.038 & 0.876 $\pm$ 0.015 & 0.838 $\pm$ 0.039 & 0.830 $\pm$ 0.040 & Base model \\ \vspace{-0.03cm}
 & 0.844 $\pm$ 0.059 & 0.904 $\pm$ 0.023 & 0.844 $\pm$ 0.059 & 0.838 $\pm$ 0.068 & Base model + Attention \\ \midrule \vspace{-0.03cm}
\multirow{6}{*}{SUR+SEC} & 0.826 $\pm$ 0.027 & 0.846 $\pm$ 0.030 & 0.826 $\pm$ 0.027 & 0.824 $\pm$ 0.029 & Base model \\ \vspace{-0.03cm}
 & \textbf{0.902 $\pm$ 0.014} &\textbf{ 0.910 $\pm$ 0.014} & \textbf{0.902 $\pm$ 0.015} & \textbf{0.902 $\pm$ 0.019} & Base model + Attention \\ \cmidrule(l){2-6}  \vspace{-0.03cm}
 & 0.828 $\pm$ 0.039 & 0.844 $\pm$ 0.036 & 0.828 $\pm$ 0.039 & 0.812 $\pm$ 0.049 & MV model (max-pooling) \\ \vspace{-0.03cm}
 & \textbf{0.966 $\pm$ 0.005} & \textbf{0.968 $\pm$ 0.008} & \textbf{0.966 $\pm$ 0.005} & \textbf{0.962 $\pm$ 0.008} & MV model (max-pooling) + Attention \\ \cmidrule(l){2-6}  \vspace{-0.03cm}
 & 0.855 $\pm$ 0.036 & 0.870 $\pm$ 0.030 & 0.850 $\pm$ 0.035 & 0.841 $\pm$ 0.040 & MV model (concatenation) \\ \vspace{-0.03cm}
 & \textbf{0.969 $\pm$ 0.004} & \textbf{0.980 $\pm$ 0.010} & \textbf{0.971 $\pm$ 0.004} & \textbf{0.970 $\pm$ 0.010} & MV model (concatenation) + Attention \\ \bottomrule \vspace{-0.03cm}
\end{tabular}
\end{table*}

\begin{figure*}[]
\centering
\includegraphics[width=0.97\linewidth]{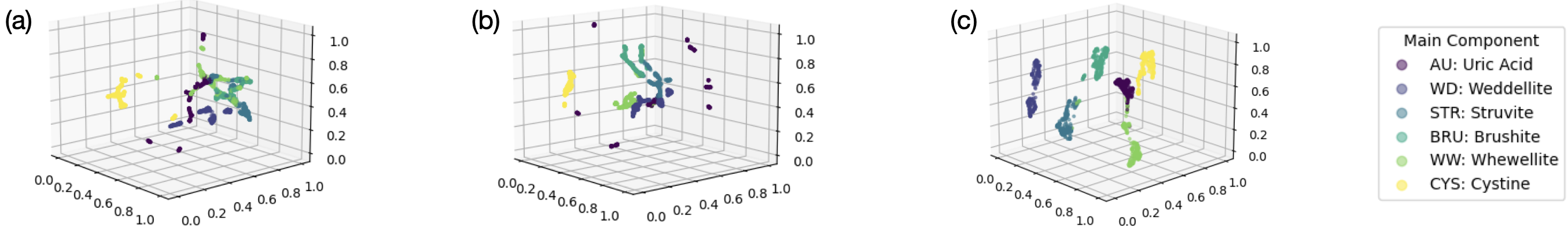}
\vspace{-0.15cm}
\caption{UMAP visualizations of the features extracted by the models. (a) No-attention mixed model (Mixed Base model), (b) Mixed Base model + Attention, and (c) MV model (max pool) + Attention. See \ref{tab:results1} for more details about the trained models.}
\label{fig:umap}
\end{figure*}

\begin{table}[]
\centering
\caption{Comparison of the performances of the models studied in this contribution (see the accuracy column in Table \ref{tab:results1}) with the model accuracy of the state-of-the-art.}
\vspace{-0.15cm}
\label{tab:results_sota}
\begin{tabular}{@{}cccc@{}}
\toprule \vspace{-0.03cm}
Method & Surface & Section & SUR+SEC \\ \midrule \vspace{-0.025cm}
Black, et al. \cite{black2020deep} & 0.735$\pm$0.190 & 0.888$\pm$0.028 & 0.801$\pm$0.138 \\\vspace{-0.025cm}
Estrade, et al. \cite{estrade2022towards} & 0.737$\pm$0.179 & 0.788$\pm$0.106 & 0.701$\pm$0.223 \\  \vspace{-0.025cm}
Lopez, et al. \cite{lopez2021assessing} &0.810$\pm$0.030  &0.880$\pm$0.023  &0.850$\pm$0.030 \\ \vspace{-0.025cm}
Lopez, et al. \cite{lopez2022boosting} & 0.832$\pm$0.012 & 0.904$\pm$0.048 & 0.856$\pm$0.001 \\ \vspace{-0.025cm}
\textbf{This proposal} & \textbf{0.888$\pm$0.028} & \textbf{0.844$\pm$0.060} & \textbf{0.966$\pm$0.005} \\ \bottomrule \vspace{-0.025cm}
\end{tabular}
\end{table}

\section{Results and discussion}
\label{res}
The three patch data (SUR, SEC and SUR+SEC) in Table \ref{tab:dataset} were separately used to assess the impact of  the attention mechanisms on the recognition of kidney stones seen in endoscopic images.
%
%
An additional experiment to assess the incorporation of attention in a MV-scheme was done to evaluate the effects of training the network on mixed data. It has been reported that the combination of different views generates valuable features for a classification using DL networks \cite{lopez2021assessing, lopez2022boosting, ochoa2022vivo}. 
In the experiment with mixed views, 9600 and 2400 patches were used for the training and the testing phases, respectively. The accuracy, precision, recall and F1-score metrics were used to assess the model performances. The results of the experiments are shown in Table \ref{tab:results1}. 
\subsection{Single-view Model}
\textbf{Surface patch results. }
As seen in Table \ref{tab:results1}, the overall accuracy after training the base model using only the weights transferred from ImageNet (i.e., without attention layers) is 0.856$\pm$0.030. 
It is noticeable in Table \ref{tab:results1} that adding attention layers to the base model increased the accuracy by 3\%, leading to value of 0.888$\pm$0.028 for this criterion. 

\textbf{Section patch results.}
The base model without attention led to an accuracy of 0.836$\pm$0.039 for section data. This accuracy reached a value of 0.844$\pm$0.059 by applying attention to the baseline model. 
This very small increase over the baseline performances is probably due to the fact that the stronger textures present in section data do less require attention layers making a focus on these features.
%
%

\textbf{Mixed (SUR+SEC) patch results. }
%
The model with attention for mixed views shows promising results (accuracy of 0.902$\pm$0.015) compared to the model without attention (0.826$\pm$0.027). An overall increase of 8.0\% is achieved for all our metrics when adding attention layers to the base model. 
Similar performance improvements were also observed in \cite{lopez2021assessing,lopez2022boosting, ochoa2022vivo} when features of both surface and section patches are used in a single training step.

The baseline + attention model achieves an increase of 4\% in terms of accuracy in comparison to \cite{lopez2022boosting}. The feature extraction part of this model was used as backbones in the MV-experiment.

\subsection{Multi-view Model}
The feature extraction layers from the models trained with the mixed dataset are duplicated and used to extract information both from the SUR and SEC views.
 The two models with the attention module and fusion strategies yielded the two best performances in this work, obtaining an overall accuracy of 0.996$\pm$0.005 and 0.969$\pm$0.004 for the max-pooling strategy and the fusion strategy, respectively.  
%
In addition, the distribution of the features by stone type also improves when attention is added.
Despite of this, it can be observed in Fig. \ref{fig:umap}.(a) and \ref{fig:umap}.(b) that the clusters (corresponding to urinary stone types) are scattered, elongated or fragmented in the three-dimensional UMAP feature space. By combining the information from different views with the MV-model and attention layers (see Fig. \ref{fig:umap}.(c)) the inter-class distances are increased, while the intra-class distances are reduced. These tighter clusters of points in the feature facilitate the classification task.

\section{Conclusions}
\label{conclusion}
The results given in this contribution demonstrate that the classification results of six different types of kidney stones can be improved by the insertion of attention mechanisms in CNN models, and that MV-schemes are also boosted by this addition. 
The experiments also show that the feature distribution by stone type is enhanced by including several attention blocks along the network as the learned features are improved, leading to larger inter-class distances and smaller intra-class distances in the feature space. 



\section*{Acknowledgments}
The authors wish to thank the AI Hub and the CIIOT at Tecnologico de Monterrey for their support for carrying the experiments reported in this paper in their NVIDIA's DGX computer.

We also to thank CONACYT for the master scholarship for Elias Alejandro Villalvazo-Avila  and the doctoral scholarship for Francisco Lopez-Tiro at Tecnologico de Monterrey.

\section*{Compliance with ethical approval}
The images were captured in medical procedures following the ethical principles outlined in the Helsinki Declaration of 1975, as revised in 2000, with the consent of the patients.

\bibliography{references.bib}{}
\bibliographystyle{IEEEtran}

\end{document}